\documentclass[11pt,a4paper]{article}
\usepackage[hyperref]{emnlp2020}
\usepackage{graphicx}
\usepackage{times}
\usepackage{latexsym}
\usepackage{caption}
\usepackage{subcaption}
\usepackage{array,multirow}
 \usepackage{float}
 \usepackage{enumitem}

\usepackage[titletoc,toc,title]{appendix}
\usepackage{hyperref}
\usepackage{microtype}

\aclfinalcopy % Uncomment this line for the final submission
\title{On Cross-Dataset Generalization in Automatic Detection of Online Abuse}

\author{Isar Nejadgholi \qquad \qquad \qquad \qquad Svetlana Kiritchenko\\
\\
National Research Council Canada \\
\texttt \footnotesize \{isar.nejadgholi,svetlana.kiritchenko\}@nrc-cnrc.gc.ca}

\date{}

\begin{document}
\maketitle
\begin{abstract}
NLP research has attained high performances in abusive language detection as a supervised classification task. While in research settings, training and test datasets are usually obtained from similar data samples, in practice systems are often applied on data that are different from the training set in topic and class distributions. 
Also, the ambiguity in class definitions inherited in this task aggravates the discrepancies between source and target datasets.
We explore the topic bias and the task formulation bias in cross-dataset generalization. 
We show that the benign examples in the Wikipedia Detox dataset are biased towards platform-specific topics. 
We identify these examples using unsupervised topic modeling and manual inspection of topics' keywords. 
Removing these topics increases cross-dataset generalization, without reducing in-domain classification performance. For a robust dataset design, we suggest applying inexpensive unsupervised methods to inspect the collected data and downsize the non-generalizable content before manually annotating for class labels. 

\end{abstract}

\section{Introduction}
\label{sec:intro}

The NLP research community has devoted significant efforts to support the safety and inclusiveness of online discussion forums by developing automatic systems to detect hurtful, derogatory or obscene utterances. 
Most of these systems are based on supervised machine learning techniques, and require annotated data. 
Several publicly available datasets have been created for the task \cite{mishra2019tackling,vidgen2020directions}. 
However, due to the ambiguities in the task definition and complexities of data collection, cross-dataset generalizability remains a challenging and under-studied issue of online abuse detection. 

Existing datasets differ in the considered types of offensive behaviour and annotation schemes, data sources and data collection methods. 
There is no agreed-upon definition of harmful online behaviour yet. 
Several terms have been used to refer to the general concept of harmful online behavior, including \textit{toxicity} \cite{hosseini2017deceiving}, \textit{hate speech} \cite{schmidt2017survey}, \textit{offensive} \cite{zampieri2019semeval} and \textit{abusive} language \cite{waseem2017understanding,vidgen2019challenges}.  
Still, in practice, every dataset only focuses on a narrow range of sub-types of such behaviours and a single online platform \cite{jurgens2019just}. 
For example, \citet{davidson2017automated} annotated tweets for three categories, \textit{Racist}, \textit{Offensive but not Racist} and \textit{Clean}, and \citet{nobata2016abusive} collected discussions from Yahoo! Finance news and applied a binary annotation scheme of \textit{Abusive} versus \textit{Clean}. 
Further, since pure random sampling usually results in small proportions of offensive examples \cite{founta2018large}, various sampling techniques are often employed. 
\citet{zampieri2019semeval} used words and phrases frequently found in offensive messages to search for potential abusive tweets. \citet{founta2018large} and \citet{razavi2010offensive} started from random sampling, then boosted the abusive part of the datasets using specific search procedures. 
\citet{hosseinmardi2015analyzing} used snowballing to collect abusive posts on Instagram. 
Due to this variability in category definitions and data collection techniques, a system trained on a particular dataset is prone to overfitting to the specific characteristics of that dataset. 
As a result, although models tend to perform well in cross-validation evaluation on one dataset, the cross-dataset generalizability remains low \cite{van2018challenges,wiegand-etal-2019-detection}. 

In this work, we investigate the impact of two types of biases originating from source data that can emerge in a cross-domain application of models: 1) task formulation bias (discrepancy in class definitions and annotation between the training and test sets) and 2) selection bias (discrepancy in the topic and class distributions between the training and test sets). 
Further, we suggest topic-based dataset pruning as a method of mitigating selection bias to increase generalizability. This approach is different from domain adaptation techniques based on data selection \cite{ruder2017learning,liu2019reinforced} in that we apply an unsupervised topic modeling method for topic discovery without using the class labels. We show that some topics are more generalizable than others. 
The topics that are specific to the training dataset lead to overfitting and, therefore, lower generalizability. 
Excluding or down-sampling instances associated with such topics before the expensive annotation step can substantially reduce the annotation costs.

We focus on the Wikipedia Detox or \textit{Wiki}-dataset, (an extension of the dataset by \citet{wulczyn2017ex}), collected from English Wikipedia talk pages and annotated for toxicity. To explore the generalizability of the models trained on this dataset, 
we create an out-of-domain test set comprising various types of abusive behaviours by combining two existing datasets, namely \textit{Waseem}-dataset \cite{waseem2016hateful} and \textit{Founta}-dataset \cite{founta2018large}, both collected from Twitter.

Our main contributions are as follows:
\begin{itemize}[leftmargin=*]

    \item We identify topics included in the \textit{Wiki}-dataset and manually examine keywords associated with the topics to heuristically determine topics' generalizability and their potential association with toxicity.  

    \item We assess the generalizability of the task formulations by training a classifier to detect the \textit{Toxic} class in the \textit{Wiki}-dataset and testing it on an out-of-domain dataset comprising various types of offensive behaviours. We find that \textit{Wiki-Toxic} is most generalizable to \textit{Founta-Abusive} and least generalizable to \textit{Waseem-Sexism}.
    
    \item We show that re-sampling techniques result in a trade-off between the True Positive and True Negative rates on the out-of-domain test set. This trade-off is mainly governed by the ratio of toxic to normal instances and not the size of the dataset. 
    
    \item  We investigate the impact of topic distribution on generalizability and show that general and identity-related topics are more generalizable than platform-specific topics.  
    
    \item We show that excluding Wikipedia-specific data instances (54\% of the dataset) does not affect the results of in-domain classification, and improves both True Positive and True Negative rates on the out-of-domain test set, unlike re-sampling methods. Through unsupervised topic modeling, such topics can be identified and excluded before annotation. The pruned version of the Wikipedia dataset is available at \url{https://github.com/IsarNejad/cross_dataset_toxicity}.

\end{itemize}

\section{Biases Originating from Source Data}

We focus on two types of biases originated from source data: task formulation and selection bias. 

\noindent \textbf{Task formulation bias:} In commercial applications, the definitions of offensive language heavily rely on community norms and context and, therefore, are imprecise, application-dependent, and constantly evolving  \cite{chandrasekharan2018internet}. 
Similarly in NLP research, despite having clear overlaps, offensive class definitions vary significantly from one study to another. 
For example, the \textit{Toxic} class in the \textit{Wiki}-dataset refers to 
aggressive or disrespectful utterances that would likely make participants leave the discussion. 
This definition of toxic language includes some aspects of racism, sexism and hateful behaviour. 
Still, as highlighted by  \citet{vidgen2019challenges},  identity-based abuse is fundamentally different from general toxic behavior. Therefore, 
the \textit{Toxic} class definition used in the \textit{Wiki}-dataset differs in its scope from the abuse-related categories as defined in the \textit{Waseem}-dataset and \textit{Founta}-dataset. 
\citet{wiegand-etal-2019-detection} converted various category sets to binary (offensive vs. normal) and demonstrated that a system trained on one dataset can identify other forms of abuse to some extent. 
We use the same methodology and examine different offensive categories in out-of-domain test sets to explore the deviation in a system's performance caused by the differences in the task definitions. \citet{fortuna-etal-2020-toxic} demonstrated that many different definitions are used for equivalent concepts, which makes most of the publicly available datasets incompatible. They suggested hierarchical multi-class annotation to formulate the online abuse detection task.

Regardless of the task formulation, abusive language can be divided into explicit and implicit \cite{waseem2017understanding}. Explicit abuse refers to utterances that include obscene and offensive expressions, such as \textit{stupid} or \textit{scum}, even though not all utterances that include obscene expressions are considered abusive in all contexts. 
Implicit abuse refers to more subtle harmful behaviours, such as stereotyping and micro-aggression. Explicit abuse is usually easier to detect by human annotators and automatic systems. Also, explicit abuse is more transferable between datasets as it is part of many definitions of online abuse, including personal attacks, hate speech, and identity-based abuse. The exact definition of implicit abuse, on the other hand, can substantially vary between task formulations as it is much dependent on the context, the author and the receiver of an utterance \cite{wiegand-etal-2019-detection}.

\noindent \textbf{Selection bias:} Selection (or sampling) bias emerge when source data, on which the model is trained, is not representative of target data, on which the model is applied \cite{shah-etal-2020-predictive}. 
We focus on two data characteristics affecting selection bias: topic distribution and class distribution. 

In practice, every dataset covers a limited number of topics, and the \textbf{topic distributions} 
depend on many factors, including the source of data, the search mechanism and the timing of the data collection. 
For example, our source dataset, \textit{Wiki}-dataset, consists of Wikipedia talk pages dating from 2004--2015. On the other hand, one of the sources of our target dataset, \textit{Waseem}-dataset, consists of tweets collected using terms and references to specific entities that frequently occur in tweets expressing hate speech. 
As a result of its sampling strategy, \textit{Waseem}-dataset includes many tweets on the topic of `women in sports'. 
\citet{wiegand-etal-2019-detection} showed that different data sampling methods result in various distributions of topics, which affects the generalizability of trained classifiers, especially in the case of implicit abuse detection. Unlike explicit abuse, implicitly abusive behaviour comes in a variety of semantic and syntactic forms. To train a generalizable classifier, one requires a training dataset that covers a broad range of topics, each with a good representation of offensive examples. 
We continue this line of work and investigate the impact of topic bias on cross-dataset generalizability by identifying and changing the distribution of topics in controlled experiments.

The amount of online abuse on mainstream platforms varies greatly but is always very low. \citet{founta2018large} found that abusive tweets form 0.1\% to 3\% of randomly collected datasets. \citet{vidgen2019much} showed that  depending on the platform the prevalence of abusive language can range between 0.001\%  and 8\%. Despite various data sampling strategies aimed at increasing the proportion of offensive instances, the \textbf{class imbalance} (the difference in class sizes) in available datasets is often severe. 
When trained on highly imbalanced data, most statistical machine learning methods exhibit a bias towards the majority class, and their performance on a minority class, usually the class of interest, suffers. 
A number of techniques have been proposed to address class imbalance in data, including data re-sampling, cost-sensitive learning, and neural network specific learning algorithms \cite{branco2016survey,haixiang2017learning,johnson2019survey}. 
In practice, simple re-sampling techniques, such as down-sampling of over-represented classes, often improve the overall performance of the classifier \cite{johnson2019survey}. However, re-sampling techniques might lead to overfitting to one of the classes causing a trade-off between True Positive and True Negative rates. When aggregated in an averaged metric such as F-score, this trade-off is usually overlooked.

\section{Datasets}

\label{sec:task formulations}
We exploit three large-scale, publicly available English datasets frequently used for the task of online abuse detection.
Our main dataset, \textit{Wiki}-dataset \cite{wulczyn2017ex}, is used as a training set. 
The out-of-domain test set is obtained by combining 
the other two datasets, \textit{Founta}-dataset \cite{founta2018large} and \textit{Waseem}-dataset \cite{waseem2016hateful}. 

\vspace{4pt}  
\noindent \textbf{Training set}: The \textit{Wiki}-dataset includes 160K comments collected from English Wikipedia discussions and annotated for \textit{Toxic} and \textit{Normal}, through crowd-sourcing\footnote{\url{https://meta.wikimedia.org/wiki/Research:Detox/Data_Release}}. Every comment is annotated by 10 workers, and the final label is obtained through majority voting. 
The class \textit{Toxic} comprises rude, hateful, aggressive, disrespectful or unreasonable comments that are likely to make a person leave a conversation\footnote{\url{https://github.com/ewulczyn/wiki-detox/blob/master/src/modeling/toxicity_question.png}}. 
The dataset consists of randomly collected comments and comments made by users blocked for violating Wikipedia’s policies to augment the proportion of toxic texts. This dataset contains  15,362 instances of \textit{Toxic} and 144,324 \textit{Normal} texts. 

\vspace{4pt}  
\noindent \textbf{Out-of-Domain test set}: The toxic portion of our test set is composed of four types of offensive language: \textit{Abusive} and \textit{Hateful} from the \textit{Founta}-dataset, and \textit{Sexist} and \textit{Racist} from the  \textit{Waseem}-dataset. For the benign examples of our test set, we use the \textit{Normal} class of the \textit{Founta}-dataset. 

The \textit{Founta}-dataset is a collection of 80K tweets crowd-annotated for four classes: \textit{Abusive}, \textit{Hateful}, \textit{Spam} and \textit{Normal}.   
The data is randomly sampled and then boosted with tweets that are likely to belong to one or more of the minority classes by deploying an iterative data exploration technique.
The \textit{Abusive} class is defined as content with any strongly impolite, rude or hurtful language that shows a debasement of someone or something, or shows intense emotions. 
The \textit{Hateful} class refers to tweets that express hatred towards a targeted individual or group, or are intended to be derogatory, to humiliate, or to insult members of a group, on the basis of attributes such as race, religion, ethnic origin, sexual orientation, disability, or gender. 
\textit{Spam} refers to posts consisted of advertising/marketing, posts selling products of adult nature, links to
malicious websites, phishing attempts and other unwanted information, usually 
sent repeatedly. Tweets that do not fall in any of the prior classes are labelled as \textit{Normal} \cite{founta2018large}. 
We do not include the \textit{Spam} class in our test set as this category does not constitute offensive language, in general. The \textit{Founta}-dataset contains 27,150 of \textit{Abusive},  4,965 of \textit{Hateful} and 53,851 of \textit{Normal} instances.

The \textit{Waseem}-dataset includes 16K manually annotated tweets, labeled as \textit{Sexist}, \textit{Racist} or \textit{Neither}. 
The corpus is collected by searching for common slurs and terms pertaining to minority groups as well as identifying tweeters that use these terms frequently. 
A tweet is annotated as \textit{Racist} or \textit{Sexist} if it uses a racial or sexist slur, attacks, seeks to silence, unjustifiably criticizes or  misrepresents a minority or defends xenophobia or sexism. Tweets that do not fall in these two classes are labeled as \textit{Neither} \cite{waseem2016hateful}.
The \textit{Neither} class represents a mixture of benign and abusive (but not sexist or racist) instances, and, therefore, is excluded from our test set.
\textit{Waseem}-dataset contains 3,430 of \textit{Sexist} and 1,976 of \textit{Racist} tweets.

\iffalse
\begin{table}[t]
\small{
\centering
\begin{tabular}{lr}
\textbf{Dataset} &\textbf{Number of instances}\\
\hline
\textit{Wiki}-dataset &\\
$\ \ \ \ $ Personal attack & 15,362\\
$\ \ \ \ $ Normal  & 144,324 \\
%$\ \ \ \ $ Total & 159,686\\ 
\textit{Founta}-dataset & \\
$\ \ \ \ $ Abusive & 27,150\\
$\ \ \ \ $ Hateful & 4,965\\
$\ \ \ \ $ Normal  &  53,851\\
%$\ \ \ \ $ Total & 85,966\\%[2pt]
\textit{Waseem}-dataset & \\
$\ \ \ \ $ Sexist  & 3,430\\
$\ \ \ \ $ Racist  & 1,976\\
%$\ \ \ \ $ Total & 5,406\\ 
\hline

\end{tabular}
\caption{Number of instances per class }
\label{tab:data_statistics}
}
\end{table}
\fi

\setlength{\tabcolsep}{3pt}
 
\begin{table}[]
\small{
\centering
\begin{tabular}{ll}
%&\small{5 terms selected from 10 top words} \\
\textbf{\small{Topics}} & \textbf{\small{Top words}} \\[2pt]
\hline 
\textbf{\small{Category 1}} & \\
$\ \ $ \small{topic 0}  & \small{know, like, thank, think, want} \\
$\ \ $ \small{topic 1}  & \small{time, like, peopl, think, life} \\[2pt]
\textbf{\small{Category 2} }& \\
$\ \ $ \small{topic 2}  & \small{suck, year, school, c*ck, p*ssi} \\
$\ \ $ \small{topic 7}   & \small{english, countri, american, nation, german} \\
$\ \ $ \small{topic 8}  & \small{kill, die, jewish, islam, israel} \\
$\ \ $ \small{topic 9}  & \small{god, christian, cast, presid, japanes} \\
$\ \ $ \small{topic 12}  & \small{person, editor, attack, accuse, user} \\
$\ \ $ \small{topic 14}  & \small{f*ck, sh*t, *ss, stupid, bastard} \\
$\ \ $ \small{topic 16}  & \small{team, footbal, gay, match, station} \\[2pt]
\textbf{\small{Category 3}} & \\
$\ \ $ \small{topic 3}  & \small{redirect, talk, categori, film, episod}\\
$\ \ $ \small{topic 4}  & \small{page, wikipedia, edit, talk, articl}\\
$\ \ $ \small{topic 5}  & \small{sourc, claim, cite, wikipedia, publish}\\
$\ \ $ \small{topic 6}  & \small{link, list, page, inform, articl}\\
$\ \ $ \small{topic 10}  & \small{delet, articl, imag, tag, copyright}\\
$\ \ $ \small{topic 11}  & \small{univers, law, scienc, theori, definit}\\
$\ \ $ \small{topic 13}  & \small{page, discuss, review, talk, templat}\\
$\ \ $ \small{topic 15}  & \small{articl, section, discuss, refer, editor}\\
$\ \ $ \small{topic 17}  & \small{http, com, www, org, wiki}\\
$\ \ $ \small{topic 18}  & \small{edit, block, vandal, user, account}\\
$\ \ $ \small{topic 19}  & \small{style, align, color, background, border}\\[2pt]
\hline
\end{tabular}
\caption{Topics identified in the \textit{Wiki}-dataset. For each topic, five of ten top words that are most representative of the assigned category are shown.
}
\label{tab:categories}
}
\end{table}
\section{Topic Analysis of the \textit{Wiki}-dataset }
\label{sec:topic categories}

We start by exploring the content of the \textit{Wiki}-dataset through topic modeling. We train a topic model using the Online Latent Dirichlet Allocation (OLDA) algorithm \cite{HoffmanLDA} as implemented in the Gensim library \cite{rehurek_lrec} with the default parameters. 
Latent Dirichlet Allocation (LDA) \cite{blei2003latent} is a Baysian probabilistic model of a collection of texts. 
Each text is assumed to be generated from a multinomial distribution over a given number of topics, and each topic is represented as a multinomial distribution over the vocabulary. 
We pre-process the texts by lemmatizing the words and removing the stop words. 
To determine the optimal number of topics, we use a coherence measure that calculates the degree of semantic similarity among the top words \cite{roder2015exploring}. Top words are defined as the most probable words to be seen conditioned on a topic. We experimented with a range of topic numbers between 10 and 30 and obtained the maximal average coherence with 20 topics. Each topic is represented by 10 top words. 
For simplicity, each text is assigned a single topic that has the highest probability. 
The full list of topics and their top words are available in the Appendix. 

We group the 20 extracted topics into three categories based on the coherency of the top words and their potential association with offensive language. 
This is done through manual examination of the 10 top words in each topic. 
Table \ref{tab:categories} shows five out of ten top words for each topic that are most representative of the assigned category. 

\vspace{4pt} 
 
\noindent\textbf{Category 1: incoherent or mixture of general topics}
    
The top words of two topics (topic 0 and topic 1) are general terms such as \textit{think}, \textit{want}, \textit{time}, and \textit{life}. This category forms 26\% of the dataset. Since these topics appear incoherent, their association with offensiveness cannot be judged heuristically. Looking at the toxicity annotations we observe that 47\% of the \textit{Toxic} comments belong to these topics. 
These comments mostly convey personal insults, usually not tied to any identity group. 
The frequently used abusive terms in these \textit{Toxic} comments include \textit{f*ck}, \textit{stupid}, \textit{idiot}, \textit{*ss}, etc.

 \vspace{4pt}  
 
\noindent\textbf{Category 2: coherent, high association with offensive language}
    
Seven of the topics can be associated with offensive language; their top words represent profanity or are related to identity groups frequently subjected to abuse. 
Topic 14 is the most explicitly offensive topic; nine out of ten top words are associated with insult and hatred. 
97\% of the instances belonging to this topic are annotated as \textit{Toxic}, with 96\% of them containing explicitly toxic words.\footnote{Following \citet{wiegand-etal-2019-detection}, we estimate the proportion of explicitly offensive instances in a dataset as the proportion of abusive instances that contain at least one word from the lexicon of abusive words by \citet{wiegand-etal-2018-inducing}.} These are generic profanities with the word \textit{f*ck} being the most frequently used word.
    
The top words of the other six topics (topics 2, 7, 8, 9, 12, and 16) include either offensive words or terms related to identity groups based on gender, ethnicity, or religion. On average, 16\% of the comments assigned to these topics are labeled as \textit{Toxic}. We manually analyzed these comments, and found that each topic (except topic 12) tends to concentrate around a specific identity group. 
Offensive comments in topic 2 mostly contain sexual slur and target female and homosexual users. 
In topic 7, comments often contain racial and ethnicity based abuse. 
Topic 8 contains physical threats, often targeting Muslims and Jewish folks (the words \textit{die} and \textit{kill} are the most frequently used content words in the offensive messages of this topic). 
Comments in topic 9 involve many terms associated with Christianity (e.g., \textit{god}, \textit{christian}, \textit{Jesus}). 
Topic 16 has the least amount of comments (0.3\% of the dataset), with the offensive messages mostly targeting gay people (the word \textit{gay} appears in 67\% of the offensive messages in this topic). 
Topic 12 is comprised of personal attacks in the context of Wikipedia admin--contributor relations. 
The most common offensive words in this topic include \textit{f*ck}, \textit{stupid}, \textit{troll}, \textit{ignorant}, \textit{hypocrite}, etc.
20\% of the whole dataset and 35\% of the comments labeled as \textit{Toxic} belong to this category.

 \vspace{4pt}  
    
\noindent\textbf{Category 3: coherent, low association with offensive language}
   
The remaining eleven topics include top words specific to Wikipedia and not directly associated with offensive language. For example, keywords of topic 4 are terms such as \textit{page}, \textit{Wikipedia}, \textit{edit} and \textit{article}, and only 0.4\% of the 10,471 instances in this topic are labeled as \textit{Toxic}. These eleven topics comprise 54\% of the comments in the dataset and 18\% of the \textit{Toxic} comments.
    
\begin{table}[]
\small{
\centering
\begin{tabular}{lrrr}
%\textbf{Behaviour Type}
\textbf{Dataset/Class}&\textbf{Cat.\#1}&\textbf{Cat.\#2}&\textbf{Cat.\#3}\\
\hline
Training Set &\\
$\ \ \ \ $ \textit{Wiki-Toxic} &48\% &34\% &18\%\\
$\ \ \ \ $ \textit{Wiki-Normal}  &24\% &18\% &58\% \\
Test Set & \\
$\ \ \ \ $ \textit{Founta-Abusive} &58\% &33\% &8\%\\
$\ \ \ \ $ \textit{Founta-Hateful} &54\% &37\% &9\%\\
$\ \ \ \ $ \textit{Waseem-Sexist}  &50\% &35\% &15\%\\
$\ \ \ \ $ \textit{Waseem-Racist}  &23\% &67\% & 10\%\\
$\ \ \ \ $ \textit{Founta-Normal}  & 51\%&28\% &21\% \\

\hline

\end{tabular}
\caption{Distribution of topic categories per class  }
\label{tab:topic distribution}
}
\end{table}

\section{Topic Distribution of the Test Set }
\label{sec:topic categories in test}
We apply the LDA topic model trained on the \textit{Wiki}-dataset as described in Section~\ref{sec:topic categories} to the Out-of-Domain test set. 
As before, each textual instance is assigned a single topic that has the highest probability. 
Table \ref{tab:topic distribution} summarizes the distribution of topics for all classes in the three datasets. 

Observe that Category 3 is the least represented category of topics across all classes, except for the \textit{Normal} class in the \textit{Wiki}-dataset. Specifically, there is a significant deviation in the topic distribution between the \textit{Wiki-Normal} and the \textit{Founta-Normal} classes. This deviation can be explained by the difference in data sources. 
Normal conversations on Twitter are more likely to be about general concepts covered in Category 1 or identity-related topics covered in Category 2 than the specific topics such as \textit{writing} and \textit{editing} in Category 3. Other than \textit{Waseem-Racist}, which has 67\% overlap with Category 2, all types of offensive behaviour in the three datasets have more overlap with the general topics (Category 1) than identity-related topics (Category 2).
For example, for the \textit{Waseem-Sexist}, 50\% of instances fall under Category 1, 35\% under Category 2 and 15\% under Category 3. 
Topic 1, which is a mixture of general topics, is the dominant topic among the \textit{Waseem-Sexist} tweets. 
Out of the topics in Category 2, most of the sexist tweets are matched to topic 2 (focused on sexism and homophobia) and topic 12 (general personal insults).

\section{Generalizability of the Model Trained on the \textit{Wiki}-dataset}
\label{sec:test}

To explore how well the \textit{Toxic} class from the \textit{Wiki}-dataset generalizes to other types of offensive behaviour, we train a binary classifier (\textit{Toxic} vs. \textit{Normal}) on the \textit{Wiki}-dataset (combining the train, development and test sets) and test it on the Out-of-Domain Test set. This classifier is expected to predict a positive (\textit{Toxic}) label for the instances of classes \textit{Founta-Abusive}, \textit{Founta-Hateful}, \textit{Waseem-Sexism} and \textit{Waseem-Racism}, and a negative (\textit{Normal}) label for the tweets in the \textit{Founta-Normal} class. We fine-tune a BERT-based classifier \cite{devlin2018bert} with a linear prediction layer, the batch size of 16 and the learning rate of $2 \times 10^{-5}$ for 2 epochs. 

\vspace{1.5mm}
\noindent \textbf{Evaluation metrics: } 
In order to investigate the trade-off between the True Positive and True Negative rates, in the following experiments we report accuracy per test class. Accuracy per class is calculated as the rate of correctly identified instances within a class. Accuracy over the toxic classes (\textit{Founta-Abusive}, \textit{Founta-Hateful}, \textit{Waseem-Sexism} and \textit{Waseem-Racism}) indicates the True Positive rate, while accuracy of the normal class (\textit{Founta-Normal}) measures the True Negative rate. 
Note that given the sizes of the positive and negative test classes, all other common metrics, such as various kinds of averaged F1-scores, can be calculated from the accuracies per class. In addition, we report macro-averaged F-score, to show the overall impact of the proposed method. 

\begin{table}[]
\small{
\centering
\begin{tabular}{lrrrr}
%\textbf{Class}
& \multicolumn{4}{c}{\textbf{\small{Test Subset}}}\\
\cline{2-5}
%& \small\textbf{\textit{Wiki}-dataset} & \small\textbf{pruned-1}& \small\textbf{pruned-2}\\ 
\textbf{Dataset/Class}&\textbf{All}&\textbf{Cat.\#1}&\textbf{Cat.\#2}&\textbf{Cat.\#3} \\
\hline
\textit{Out-of-Domain} - Toxic & \\
$\ \ \ \ $ \textit{Founta-Abusive} &0.94 &0.94 &\textbf{0.96}&0.91\\
$\ \ \ \ $ \textit{Founta-Hateful} &0.62 &\textbf{0.65} &0.62&0.43\\
$\ \ \ \ $ \textit{Waseem-Sexist}  &0.26 &\textbf{0.29} &0.26&0.17\\
$\ \ \ \ $ \textit{Waseem-Racist}  &0.35 &\textbf{0.37} & 0.36&0.20\\

\textit{Out-of-domain} - Normal & \\
$\ \ \ \ $ \textit{Founta-Normal}  & 0.96&0.95 &0.97&\textbf{0.99} \\
\hline

\end{tabular}
\caption{Accuracy per test class and topic category for a classifier trained on \textit{Wiki}-dataset. Best results in each row are in bold.}
\label{tab:accuracies}
}
\end{table}

\vspace{1.5mm}
\noindent \textbf{Results:} The overall performance of the classifier on the Out-of-Domain test set is quite high: macro-averaged $F_{1} = 0.90$. 
However, when the test set is broken down into the 20 topics of the \textit{Wiki}-dataset and the accuracy is measured within the topics, the results vary greatly. For example, for the instances that fall under topic 14, the explicitly offensive topic, the F1-score is 0.99. For topic 15, a Wikipedia-specific topic, the F1-score is 0.80. 
Table~\ref{tab:accuracies} shows the overall accuracies for each test class as well as the accuracies for each topic category (described in Section \ref{sec:topic categories}) within each class.

For the class\textit{ Founta-Abusive}, the classifier 
achieves 94\% accuracy. 
12\% of the \textit{Founta-Abusive} tweets fall under the explicitly offensive topic (topic 14), and those tweets are classified with a 100\% accuracy. The accuracy score is highest on Category 2 and lowest on Category 3. 
For the \textit{Founta-Hateful} class, the classifier recognizes 62\% of the tweets correctly. The accuracy score is highest on Category 1 and lowest on Category 3. 8\% of the \textit{Founta-Hateful} tweets fall under the explicitly offensive topic (topic 14), and are classified with a 99\% accuracy. 
For the \textit{Founta-Normal} class, the classifier recognizes 96\% of the tweets correctly. Unlike the \textit{Founta-Abusive} and \textit{Founta-Hateful} class, for the \textit{Founta-Normal} class, the highest accuracy is achieved on Category 3. 0.1\% of the \textit{Founta-Normal} tweets fall under the explicitly offensive topic, and only 26\% of them are classified correctly.

The accuracy of the classifier on the \textit{Waseem-Sexist} and \textit{Waseem-Racist} classes is 0.26 and 0.35, respectively.
This indicates that the \textit{Wiki}-dataset, annotated for toxicity, is not well suited for detecting sexist or racist tweets. This observation could be explained by the fact that none of the coherent topics extracted from the \textit{Wiki}-dataset is associated strongly with sexism or racism. Nevertheless, the tweets that fall under the explicit abuse topic (topic 14) are recognized with a 100\% accuracy. 
Topic 8, which contains abuse mostly directed towards Jewish and Muslim people, is the most dominant topic in the \textit{Racist} class (32\% of the class) 
and the accuracy score on this topic is the highest, 
after the explicitly offensive topic.  
The \textit{Racist} class overlaps the least with Category 3 (see Table \ref{tab:topic distribution}), and the lowest accuracy score is obtained on this category. The definitions of the \textit{Toxic} and \textit{Racist} classes overlap mostly in general and identity-related abuse, therefore higher accuracy scores are obtained in Categories 1 and 2. Similar to \textit{Racist} tweets, \textit{Sexist} tweets have the least overlap and the lowest accuracy score on Category 3. The accuracy score is the highest on the explicitly offensive topic (100\%) and varies substantially across other topics.

\subsection{Discussion}

The generalizability of the classifier trained on the \textit{Wiki}-dataset is affected by at least two factors: task formulation and topic distributions.

\noindent \textbf{The impact of task formulation:} 
From task formulations described in Section \ref{sec:task formulations}, observe that the \textit{Wiki}-dataset defines the class  \textit{Toxic} in a general way. The class \textit{Founta-Abusive} is also a general formulation of offensive behaviour. The similarity of these two definitions is reflected clearly in our results. The classifier trained on the \textit{Wiki}-dataset reaches 96\% accuracy  on the \textit{Founta-Abusive} class. Unlike the \textit{Founta-Abusive} class, the other three labels included in our analysis formulate a specific type of harassment against certain targets. Our topic analysis of the \textit{Wiki}-dataset reveals that this dataset includes profanity and hateful content directed towards minority groups but the dataset is extremely unbalanced in covering these topics. Therefore, not only is the number of useful examples for learning these classes small, but the classification models do not learn these classes effectively because of the skewness of the training dataset. This observation is in line with the fact that the trained classifier detects some of the \textit{Waseem-Racist}, \textit{Waseem-Sexist} and \textit{Founta-Hateful} tweets correctly, but overall performs poorly on these classes.

 \begin{figure}
\centering
\includegraphics[width= 7cm]{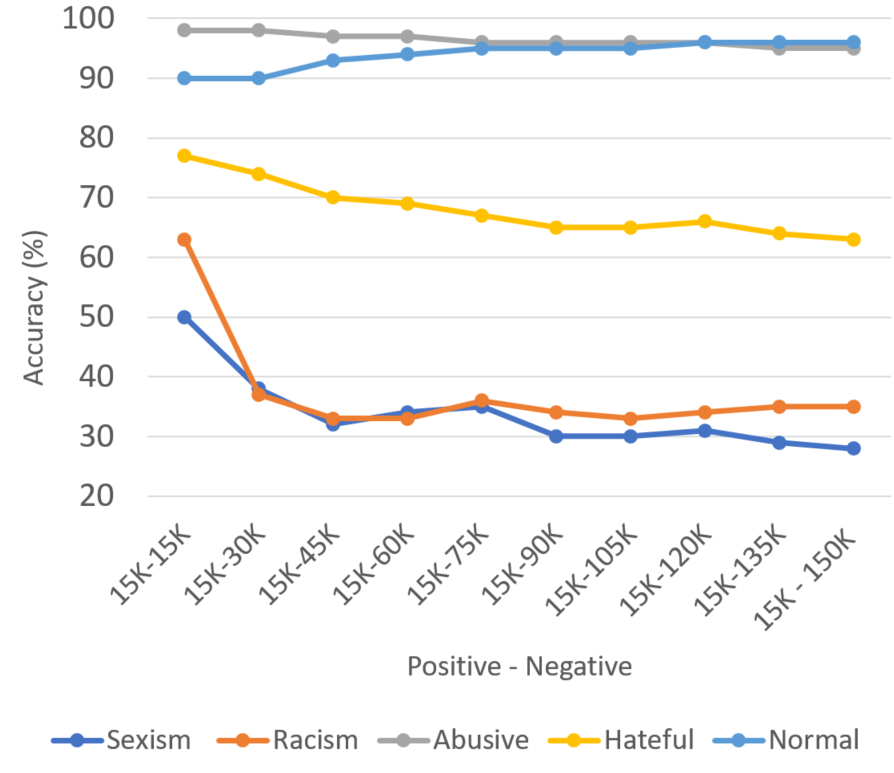}
\caption{The classifier's performance on various classes when trained on subsets of the \textit{Wiki}-dataset with specific class distributions.}
\label{fig:balance effect}
\end{figure}

\noindent \textbf{The impact of topic distribution:} 
Our analysis shows that independent of the class labels, for all the abuse-related test classes, the trained classifier performs worst when test examples fall under Category 3. Intuitively, this means that the platform-specific topics with low association with offensive language are least generalizable in terms of learning offensive behaviour. Categories 1 and 2, which include a mixture of general and identity-related topics with high potential for offensiveness, have more commonalities across datasets.

\section{Impact of Data Size, Class and Topic Distribution on Generalizability}
\label{sec:quantitative}
Our goal is to measure the impact of various topics on generalization. However, modifying the topic distribution will impact the class distribution and data size. To control for this, we first analyze the impact of class distribution and  data size on the classifier's performance. Then, we study the effect of topic distribution by limiting the training data to different topic categories.

\noindent \textbf{Impact of class distribution:} 
The class distribution in the \textit{Wiki}-dataset is fairly imbalanced; the ratio of the size of \textit{Wiki-Toxic} to \textit{Wiki-Normal} is 1:10. 
Class imbalance can lead to poor predictive performance on minority classes, as most of the learning algorithms are developed with the assumption of the balanced class distribution. To investigate the impact of the class distribution on generalization, we keep all the \textit{Wiki-Toxic} instances and randomly sample the \textit{Wiki-Normal} class to build the training sets with various ratios of toxic to normal instances.

Figure \ref {fig:balance effect} shows the classifier's accuracy on the test classes when trained on subsets with different class distributions. Observe that with the increase of the \textit{Wiki-Normal} class size in the training dataset, the accuracy on all offensive test classes decreases while the accuracy on the \textit{Founta-Normal} class increases. 
The classifier assigns more instances to the the \textit{Normal} class resulting in a lower True Positive (accuracy on the offensive classes) and a higher True Negative (accuracy on the \textit{Normal} class) rates. The drop in accuracy is significant for the  \textit{Waseem-Sexist}, \textit{Waseem-Racist} and \textit{Waseem-Hateful} classes and relatively minor for the \textit{Founta-Abusive} class. 
Note that the impact of the class distribution is not reflected in the overall F1-score. The classifier trained on a balanced data subset (with class size ratio of 1:1) reaches 0.896 macro-averaged F1-score, which is very close to the F1-score of 0.899 resulted from training on the full dataset with the 1:10 class size ratio. However, in practice, the designers of such systems need to decide on the preferred class distribution depending on the distribution of classes in the test environment and the significance of the consequences of the False Positive and False Negative outcomes.  

\begin{figure}
\centering
\includegraphics[width= 0.9\linewidth]{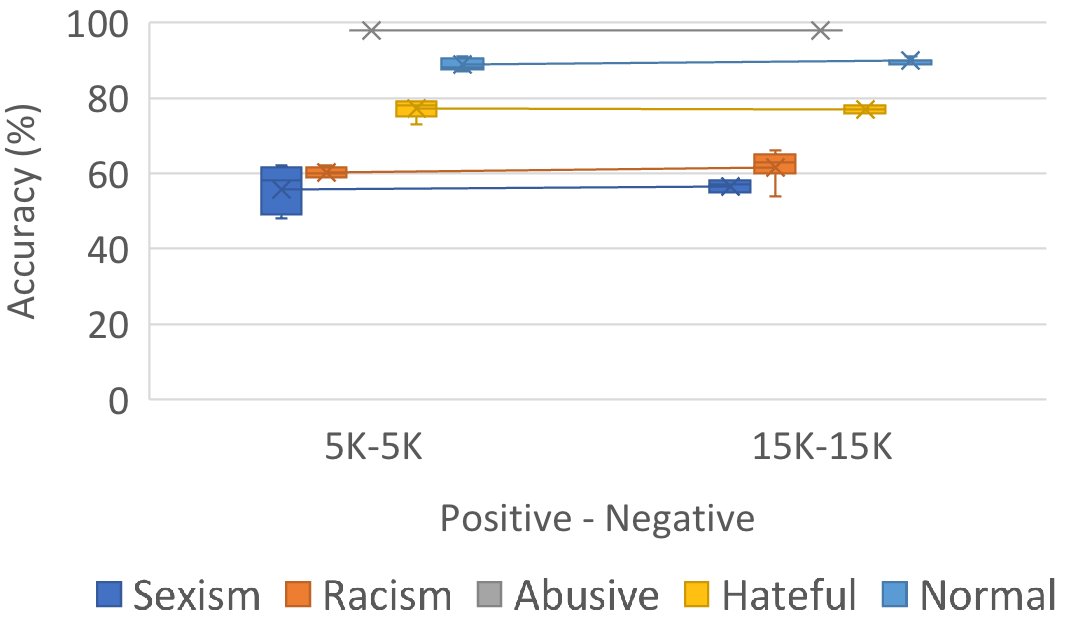}
\caption{The classifier's average performance on various classes when trained on balanced subsets of the \textit{Wiki}-dataset of different sizes.}
\label{fig:size effect}
\end{figure}

\noindent \textbf{Impact of dataset size: } To investigate the impact of the size of the training set, we fix the class ratio at 1:1 and compare the classifier's performance when trained on data subsets of different sizes. 
We randomly select subsets from the \textit{Wiki}-dataset with sizes of 10K (5K \textit{Toxic} and 5K \textit{Normal} instances) and 30K (15K \textit{Toxic} and 15K \textit{Normal} instances). Each experiment is repeated 5 times, and the averaged results are presented in Figure \ref{fig:size effect}. The height of the box shows the standard deviation of accuracies. Observe that the average accuracies remain unchanged when the dataset's size triples at the same class balance ratio. This finding contrasts with the general assumption that more training data results in a higher classification performance.

\begin{figure}[]
\centering
\begin{subfigure}{0.45\textwidth}
  \centering
  \includegraphics[width = 7cm]{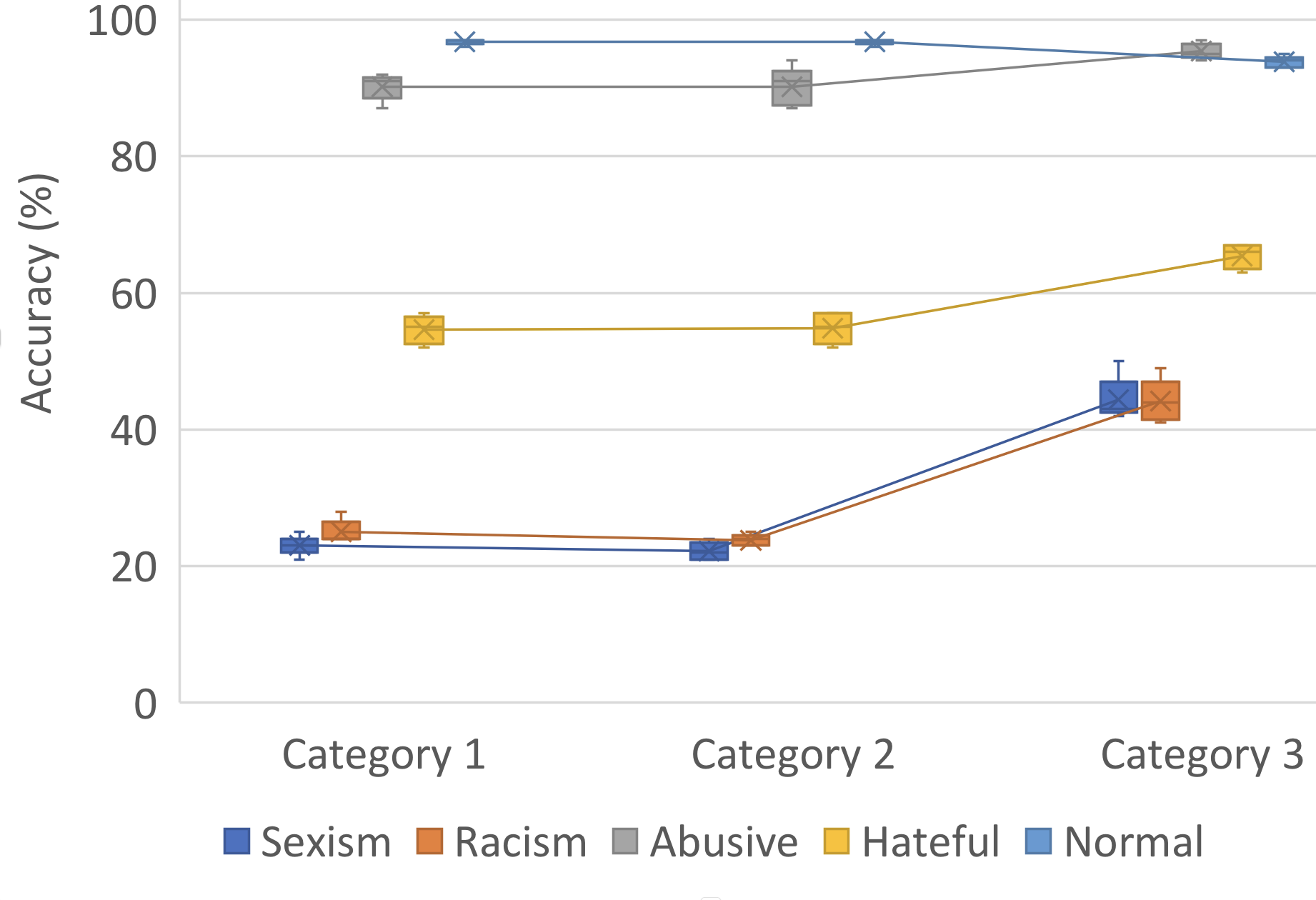}
  \caption{toxic - normal : 3K - 27K }
  \label{fig:3K-3K}
\end{subfigure}
\begin{subfigure}{0.45\textwidth}
  \centering
  \includegraphics[width = 7cm]{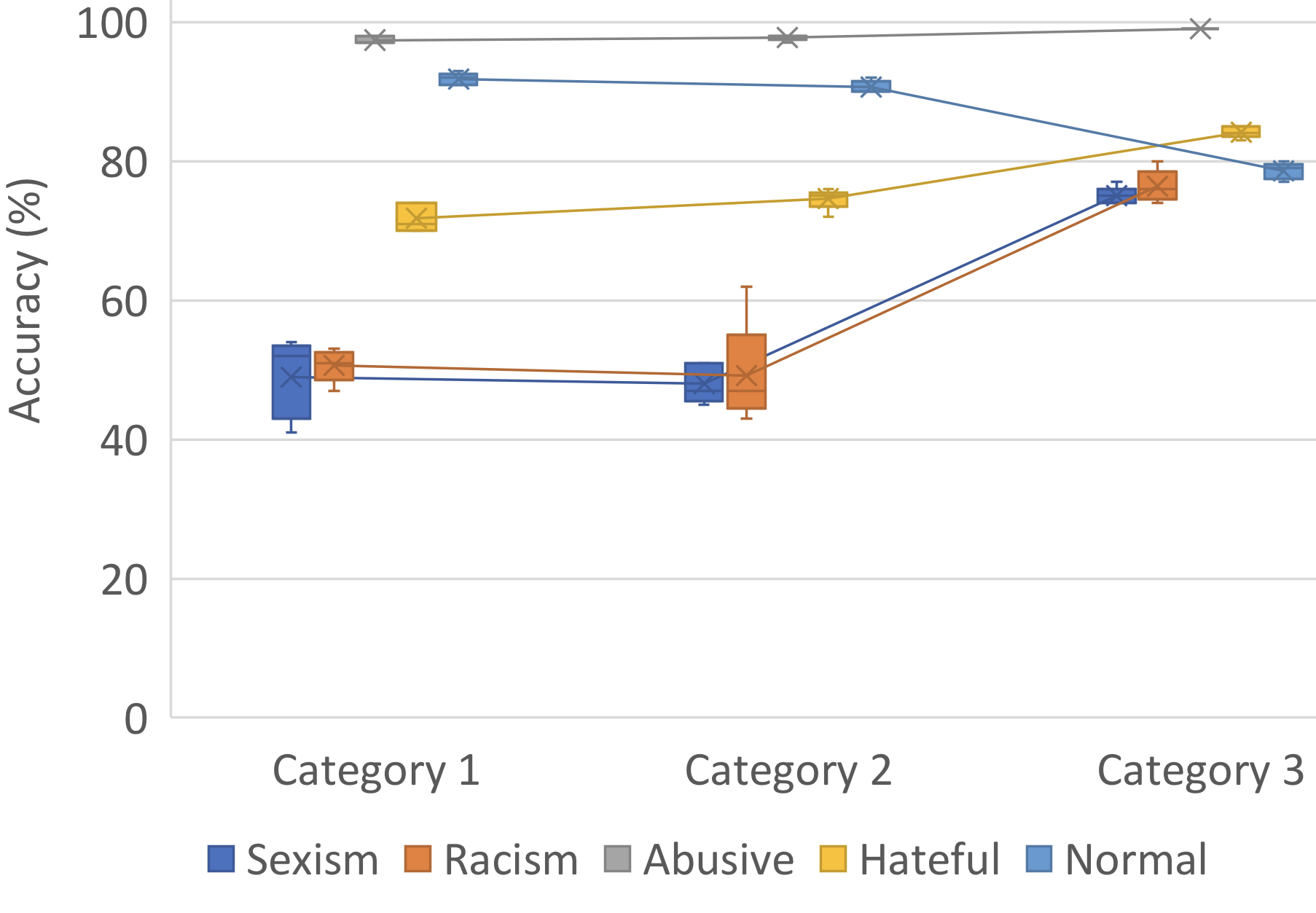}
  \caption{toxic - normal : 3K - 3K }
  \label{fig:3K-27K}
\end{subfigure}
\caption{The classifier's performance on various classes when trained on specific topic categories.}
\label{fig:topic effect}
\end{figure}

\noindent \textbf{Impact of topics: }In order to measure the impact of topics covered in the training dataset, we compare the classifier's performance when trained on only one of the three categories of topics described in Section \ref{sec:topic categories}. To control for the effect of class balance and dataset size, we run the experiments for two cases of toxic-to-normal ratios, 3K-3K and 3K-27K. Each experiment is repeated 5 times, and the average accuracy per class is reported in Figure \ref{fig:topic effect}. 

For both cases of class size ratios, shown in Figures \ref{fig:3K-3K} and \ref{fig:3K-27K}, we notice that the classifier trained on instances belonging to Category 3 reaches higher accuracies  on the offensive classes, but a significantly lower accuracy on the \textit{Founta-Normal} class. 
The benign part of Category 3 is overwhelmed by Wikipedia-specific examples. Therefore, utterances dissimilar to these topics are labelled as \textit{Toxic}, leading to a high accuracy on the toxic classes and a low accuracy on the \textit{Normal} class. This is an example of the negative impact of topic bias on the detection of offensive utterances.

In contrast, the classifiers trained on Categories 1 and 2 perform comparably across test classes. The classifier trained on Category 2 is slightly more effective in recognizing \textit{Founta-Hateful} utterances, especially when the training set is balanced. This observation can be explained by a better representation of identity-related hatred in Category 2.

\section{Removing Platform-Specific Instances from the Training Set }

We showed that a classifier trained on instances from Category 3 suffers a big loss in accuracy on the \textit{Normal} class. 
Here, we investigate how the performance of a classifier trained on the full \textit{Wiki}-dataset changes when the Category 3 instances (all or the benign part only) are removed from the training set. Table~\ref{tab:exclude benign impact} shows the results. Observe that removing the domain-specific benign examples, referred to as `excl. C3 \textit{Normal}' in Table~\ref{tab:exclude benign impact}, improves the accuracies for all classes. 
As demonstrated in the previous experiments, this improvement cannot be attributed to the changes in the class balance ratio or the size of the training set, as both these factors cause a trade-off between True Positive and True Negative rates. Removing the Wikipedia-specific topics from the \textit{Wiki}-dataset mitigates the topic bias and leads to this improvement. 
  
Similarly, when all the instances of Category 3 are removed from the training set (`excl. C3 all' in Table~\ref{tab:exclude benign impact}), the accuracy does not suffer and actually slightly improves on all classes, except \textit{Waseem-Racist}. This is despite the fact that the training set has 58\% less instances in the \textit{Normal} class and 18\% less instances in the \textit{Toxic} class.  
The overall macro-averaged F1-score on the full Out-of-Domain test set also slightly improves when the instances of Category 3 are excluded from the training data (Table~\ref{tab:F-score}). 
Removing all the instances of Category 3 is particularly interesting since it can be done only with inspection of topics and without using the class labels. 

To assess the impact of removing Wikipedia-specific examples on in-domain classification, we train a model on the training set of the \textit{Wiki}-dataset, with and without excluding Category 3 instances, and evaluate it on the full test set of the \textit{Wiki}-dataset. We observe that the in-domain performance does not suffer from removing Category 3 from the training data (Table \ref{tab:F-score}).

\setlength{\tabcolsep}{7pt}
 
\begin{table}[]
\small{
\centering
\begin{tabular}{lccc}
&\multicolumn{3}{c}{\textbf{\small{Training Set}}}\\
\cline{2-4}
 
\small\textbf{Dataset/Class}& \small\textbf{\textit{Wiki}} & \small\textbf{\textit{Wiki} excl.}& \small\textbf{\textit{Wiki} excl.}\\ 
&  & \small\textbf{C3 \textit{Normal}}& \small\textbf{C3 all}\\ 

\hline

Toxic & \\
$\ \ \ \ $ \textit{\small{Founta-Abusive}}  & 0.94 &0.96&0.95\\
$\ \ \ \ $ \textit{\small{Founta-Hateful}}  & 0.62 &0.67&0.65\\
$\ \ \ \ $ \textit{\small{Waseem-Sexist}}  & 0.26  &0.30 &0.28\\
$\ \ \ \ $ \textit{\small{Waseem-Racist}} & 0.35 &0.40 &0.32\\

Normal & \\
$\ \ \ \ $ \textit{\small{Founta-Normal}}  & 0.96 &0.97 &0.97\\

\hline
\end{tabular}
\caption{Accuracy per Out-of-Domain test class for a classifier trained on the \textit{Wiki}-dataset, and the \textit{Wiki}-dataset with Category 3 instances (\textit{Normal} only or all) excluded.}
\label{tab:exclude benign impact}
}
\end{table}

\section{Discussion }
In the task of online abuse detection, both False Positive and False Negative errors can lead to significant harm as one threatens the freedom of speech and ruins people's reputations, and the other ignores hurtful behaviour. Although balancing the class sizes has been traditionally exploited when dealing with imbalanced datasets, we showed that balanced class sizes may lead to high misclassification of normal utterances while improving the True Positive rates. This trade-off is not necessarily reflected in aggregated evaluation metrics such as F1-score but has important implications in real-life applications. We suggest evaluating each class (both positive and negative) separately taking into account the potential costs of different types of errors. Furthermore, our analysis reveals that for generalizability, the size of the dataset is not as important as the class and topic distributions.

We analyzed the impact of the topics included in the \textit{Wiki}-dataset and showed that mitigating the topic bias improves accuracy rates across all the out-of-domain positive and negative classes. Our results suggest that the sheer amount of normal comments included in the training datasets might not be necessary and can even be harmful for generalization if the topic distribution of normal topics is skewed. When the classifier is trained on Category~3 instances only (Figure \ref{fig:topic effect}), the \textit{Normal} class is attributed to the over-represented topics, leading to high misclassification of normal texts or high False Positive rates.

In general, when collecting new datasets, texts can be inspected through topic modeling using simple heuristics (e.g., keep topics related to demographic groups often subjected to abuse) in an attempt to balance the distribution of various topics and possibly sub-sample over-represented and less generalizable topics (e.g., high volumes of messages related to an incident with a celebrity figure happened during the data collection time) before the expensive annotation step.

\setlength{\tabcolsep}{8pt}
 
\begin{table}[]
\small{
\centering
\begin{tabular}{lccc}
&\multicolumn{3}{c}{\textbf{\small{Training Set}}}\\
\cline{2-4}

\small\textbf{Test Set}& \small\textbf{\textit{Wiki}} & \small\textbf{\textit{Wiki} excl.}& \small\textbf{\textit{Wiki} excl.}\\ 
&  & \small\textbf{C3 \textit{Normal}}& \small\textbf{C3 all}\\ 

\hline

\textit{Out-of-Domain} & 0.89 &0.91 &0.90\\
\textit{In-Domain} & 0.89&0.89&0.89\\

\hline
\end{tabular}
\caption{Macro-averaged F1-score for a classifier trained on portions of the \textit{Wiki}-dataset and evaluated on the in-domain and out-of-domain test sets. }
\label{tab:F-score}
}
\end{table}

\section{Conclusion }

Our work highlights the importance of heuristic scrutinizing of topics in collected datasets before performing a laborious and expensive annotation. We suggest that unsupervised topic modeling and manual assessment of extracted topics can be used to mitigate the topic bias.  
In the case of the \textit{Wiki}-dataset, we showed that more than half of the dataset can be safely removed without affecting either the in-domain or the out-of-domain performance. 
For future work, we recommend that topic analysis, augmentation of topics associated with offensive vocabulary and targeted demographics, and filtering of non-generalizable topics should be applied iteratively during data collection.  

\bibliography{main}
\bibliographystyle{acl_natbib}

\begin{appendices}
\section{Topics in \textit{Wiki}-dataset}
\noindent The following 20 topics are extracted from \textit{Wiki}-dataset using LDA algorithm. Each topic is represented as a multinomial combination of 10 keywords. The number of utterances that fall under each topic and the percentage of toxic utterances is reported. 

\vspace{5pt}
\noindent\textbf{Category 1: incoherent and mixture of general topics}

\vspace{5pt}
\noindent--------------------------------\\
Topic \#0: \\
0.025*"know" + 0.019*"like" + 0.019*"thank" + 0.015*"think" + 0.014*"want" + 0.013*"look" + 0.012*"ll" + 0.012*"ve" + 0.012*"hi" + 0.011*"time"\\\\
31210 documents  - 0.195 of dataset\\ 
0.18\% labeled as \textit{Toxic} and 36\% of all \textit{Toxic}s\\
--------------------------------\\
Topic \#1: \\
0.012*"time" + 0.010*"like" + 0.009*"peopl" + 0.009*"think" + 0.007*"life" + 0.007*"year" + 0.007*"idiot" + 0.007*"day" + 0.006*"know" + 0.006*"drink"\\\\
9598 documents  - 0.060 of dataset\\
0.19\% labeled as \textit{Toxic} and 12\% of all \textit{Toxic}s\\
--------------------------------\\
\noindent\textbf{Category 2: coherent and high association with offensive language}

\noindent Topic \#2: \\
0.020*"suck" + 0.015*"year" + 0.010*"new" + 0.009*"citi" + 0.009*"school" + 0.008*"cock" + 0.008*"old" + 0.008*"dick" + 0.007*"pussi" + 0.007*"women"\\\\
6466 documents  - 0.040 of all the documnets\\
0.18\% labeled as \textit{Toxic} and 8\% of all \textit{Toxic}s\\
--------------------------------\\
Topic \#7: \\
0.018*"english" + 0.017*"languag" + 0.016*"peopl" + 0.011*"countri" + 0.010*"american" + 0.009*"nation" + 0.008*"term" + 0.008*"use" + 0.008*"word" + 0.007*"german"\\\\
7706 documents  - 0.048 of dataset\\
0.06\% labeled as \textit{Toxic} and 3\% of all \textit{Toxic}s\\
--------------------------------\\
Topic \#8: \\
0.047*"kill" + 0.038*"live" + 0.027*"pro" + 0.024*"die" + 0.023*"eat" + 0.018*"jewish" + 0.017*"anti" + 0.017*"islam" + 0.016*"al" + 0.016*"israel"\\\\
997 documents  - 0.006 of dataset\\
0.34\% labeled as \textit{Toxic} and 2\% of all \textit{Toxic}s\\
--------------------------------\\
Topic \#9: \\
0.024*"god" + 0.020*"book" + 0.019*"christian" + 0.009*"jesus" + 0.009*"cast" + 0.008*"king" + 0.008*"prime" + 0.008*"william" + 0.008*"presid" + 0.008*"japanes"\\\\
1102 documents  - 0.007 of dataset\\
0.12\% labeled as \textit{Toxic} and 1\% of all \textit{Toxic}s\\
--------------------------------\\
Topic \#12: \\
0.014*"person" + 0.013*"editor" + 0.013*"admin" + 0.011*"attack" + 0.011*"say" + 0.010*"peopl" + 0.009*"like" + 0.009*"wikipedia" + 0.008*"accus" + 0.008*"user"\\\\
13949 documents  - 0.087 of dataset\\
0.13\% labeled as \textit{Toxic} and 12\% of all \textit{Toxic}s\\
--------------------------------\\
Topic \#14: \\
0.187*"fuck" + 0.078*"shit" + 0.056*"ass" + 0.051*"stupid" + 0.045*"bastard" + 0.037*"em" + 0.033*"bitch" + 0.032*"moron" + 0.030*"cunt" + 0.027*"hate"\\\\
1294 documents  - 0.008 of dataset\\
0.97\% labeled as \textit{Toxic} and 8\% of all \textit{Toxic}s\\
--------------------------------\\
Topic \#16: \\
0.040*"team" + 0.031*"footbal" + 0.029*"infobox" + 0.027*"award" + 0.023*"win" + 0.022*"gay" + 0.015*"air" + 0.015*"engin" + 0.015*"match" + 0.014*"station"\\\\
475 documents  - 0.003 of dataset\\
0.13\% labeled as \textit{Toxic} and 0.0\% of all \textit{Toxic}s\\
--------------------------------\\

\noindent\textbf{Category 3: coherent and low association with offensive language}

\noindent Topic \#3: \\
0.047*"redirect" + 0.040*"talk" + 0.039*"utc" + 0.036*"categori" + 0.032*"film" + 0.017*"episod" + 0.013*"merg" + 0.012*"octob" + 0.012*"decemb" + 0.011*"januari"\\
3023 documents  - 0.019 of dataset\\
0.04\% labeled as \textit{Toxic} and 1\% of all \textit{Toxic}s\\
--------------------------------\\
Topic \#4: \\
0.086*"page" + 0.082*"wikipedia" + 0.059*"edit" + 0.043*"talk" + 0.035*"help" + 0.033*"articl" + 0.028*"thank" + 0.022*"question" + 0.016*"ask" + 0.015*"revert"\\
10097 documents  - 0.063 of dataset\\
0.04\% labeled as \textit{Toxic} and 3\% of all \textit{Toxic}s\\
--------------------------------\\
Topic \#5: \\
0.086*"sourc" + 0.021*"reliabl" + 0.015*"claim" + 0.012*"cite" + 0.012*"refer" + 0.012*"wikipedia" + 0.011*"inform" + 0.011*"fact" + 0.010*"publish" + 0.010*"research"\\\\
9439 documents  - 0.059 of dataset\\
0.04\% labeled as \textit{Toxic} and 3\% of all \textit{Toxic}s\\
--------------------------------\\
Topic \#6: \\
0.034*"link" + 0.029*"list" + 0.023*"add" + 0.022*"page" + 0.014*"game" + 0.013*"inform" + 0.011*"articl" + 0.010*"date" + 0.010*"chang" + 0.009*"googl"\\\\
7807 documents  - 0.049 of dataset\\
0.03\% labeled as \textit{Toxic} and 1\% of all \textit{Toxic}s\\
--------------------------------\\
Topic \#10: \\
0.102*"delet" + 0.048*"articl" + 0.045*"imag" + 0.033*"wikipedia" + 0.029*"tag" + 0.028*"copyright" + 0.025*"file" + 0.025*"notabl" + 0.024*"page" + 0.017*"use"\\\\
6308 documents  - 0.040 of dataset\\
0.02\% labeled as \textit{Toxic} and 1\% of all \textit{Toxic}s\\
--------------------------------\\
Topic \#11: \\
0.017*"univers" + 0.017*"th" + 0.012*"law" + 0.012*"scienc" + 0.011*"theori" + 0.009*"capit" + 0.008*"centuri" + 0.008*"definit" + 0.008*"state" + 0.007*"student"\\\\
1875 documents  - 0.012 of dataset\\
0.03\% labeled as \textit{Toxic} and 0.0\% of all \textit{Toxic}s\\
--------------------------------\\
Topic \#13: \\
0.028*"page" + 0.025*"discuss" + 0.024*"review" + 0.024*"talk" + 0.023*"thank" + 0.023*"request" + 0.020*"vertic" + 0.019*"comment" + 0.018*"templat" + 0.017*"wp"\\\\
7555 documents  - 0.047 of dataset\\
0.02\% labeled as \textit{Toxic} and 1\% of all \textit{Toxic}s\\
--------------------------------\\
Topic \#15: \\ 
0.050*"articl" + 0.011*"think" + 0.011*"section" + 0.009*"wp" + 0.009*"discuss" + 0.008*"refer" + 0.008*"editor" + 0.008*"point" + 0.008*"need" + 0.007*"chang"\\\\
30128 documents  - 0.189 of dataset\\
0.02\% labeled as \textit{Toxic} and 3\% of all \textit{Toxic}s\\
--------------------------------\\
Topic \#17: \\
0.054*"http" + 0.045*"com" + 0.033*"www" + 0.033*"org" + 0.026*"en" + 0.017*"state" + 0.015*"wiki" + 0.013*"unit" + 0.011*"compani" + 0.008*"html"\\\\
2757 documents  - 0.017 of dataset\\
0.04\% labeled as \textit{Toxic} and 1\% of all \textit{Toxic}s\\
--------------------------------\\
Topic \#18: \\
0.121*"edit" + 0.097*"block" + 0.038*"vandal" + 0.030*"user" + 0.029*"account" + 0.028*"stop" + 0.027*"ip" + 0.023*"war" + 0.022*"page" + 0.021*"revert"\\\\
6559 documents  - 0.041 of dataset\\
0.09\% labeled as \textit{Toxic} and 4\% of all \textit{Toxic}s\\
--------------------------------\\
Topic \#19: \\
0.130*"style" + 0.096*"px" + 0.069*"align" + 0.059*"color" + 0.052*"background" + 0.051*"pad" + 0.044*"middl" + 0.038*"border" + 0.033*"solid" + 0.023*"size"\\\\
1341 documents  - 0.008 of dataset\\
0.06\% labeled as \textit{Toxic} and 1\% of all  \textit{Toxic}s\\
\end{appendices}
\end{document}